\documentclass[letterpaper, 10 pt, conference]{ieeeconf}  

\IEEEoverridecommandlockouts                              

\usepackage{amsmath}
\usepackage{amssymb}
\usepackage{tikz}
\usepackage[capitalise]{cleveref}
\usepackage{booktabs, multirow} 
\usepackage{subcaption}

\usepackage{url}

\usepackage{fancyhdr}
\fancypagestyle{withfooter}{
  
  \fancyhead[L]{}
  \fancyhead[R]{}
  \fancyfoot[C]{\footnotesize Presented at the 2025 IEEE ICRA Workshop on Field Robotics}
}

\title{Evaluating Robustness of Deep Reinforcement Learning for Autonomous Surface Vehicle Control in Field Tests}

\author{Luis F. W. Batista$^{1,2}$, Stéphanie Aravecchia$^{2}$, Seth Hutchinson$^{1}$, and Cedric Pradalier$^{2}$
\thanks{$^{1}$ are with Georgia Institute of Technology, Atlanta, USA}%
\thanks{$^{2}$ are with GeorgiaTech Europe - IRL2958 GT-CNRS, Metz, France}%
}

\begin{document}

\maketitle
\thispagestyle{withfooter}
\pagestyle{withfooter}

\begin{abstract}
Despite significant advancements in Deep Reinforcement Learning (DRL) for Autonomous Surface Vehicles (ASVs), their robustness in real-world conditions, particularly under external disturbances, remains insufficiently explored. In this paper, we evaluate the resilience of a DRL-based agent designed to capture floating waste under various perturbations.
We train the agent using domain randomization and evaluate its performance in real-world field tests, assessing its ability to handle unexpected disturbances such as asymmetric drag and an off-center payload. We assess the agent’s performance under these perturbations in both simulation and real-world experiments, quantifying performance degradation and benchmarking it against an MPC baseline. Results indicate that the DRL agent performs reliably despite significant disturbances.
Along with the open-source release of our implementation\footnote{\url{https://github.com/luisfelipewb/RL4WasteCapture/}}, 
we provide insights into effective training strategies, real-world challenges, and practical considerations for deploying DRL-based ASV controllers.
\end{abstract}

\section{Introduction}

Plastic pollution in water bodies poses a significant threat to global sustainability, with severe consequences for marine life, ecosystems, and human health \cite{van2020plastic}. Tackling this issue requires not only preventing further pollution but also mitigating the damage already caused. Autonomous Surface Vehicles (ASVs) have increasingly gained recognition for their potential in water quality monitoring and waste removal \cite{FornaiwaterSampling, chang2021autonomous}. Autonomous cleaning tasks using ASVs have attracted attention from both academic \cite{zhou2021time} and industrial \cite{orca_uboat} research.

\begin{figure}[tb]
    \centering
    \begin{subfigure}{0.49\linewidth}
        \includegraphics[width=\linewidth]{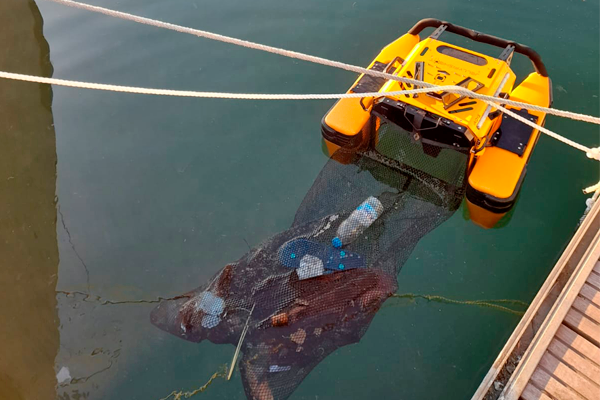}
        \caption{Jellyfishbot from IADYS}
        \label{fig:jellyfishbot}
    \end{subfigure}
    \begin{subfigure}{0.49\linewidth}
        \includegraphics[width=\linewidth]{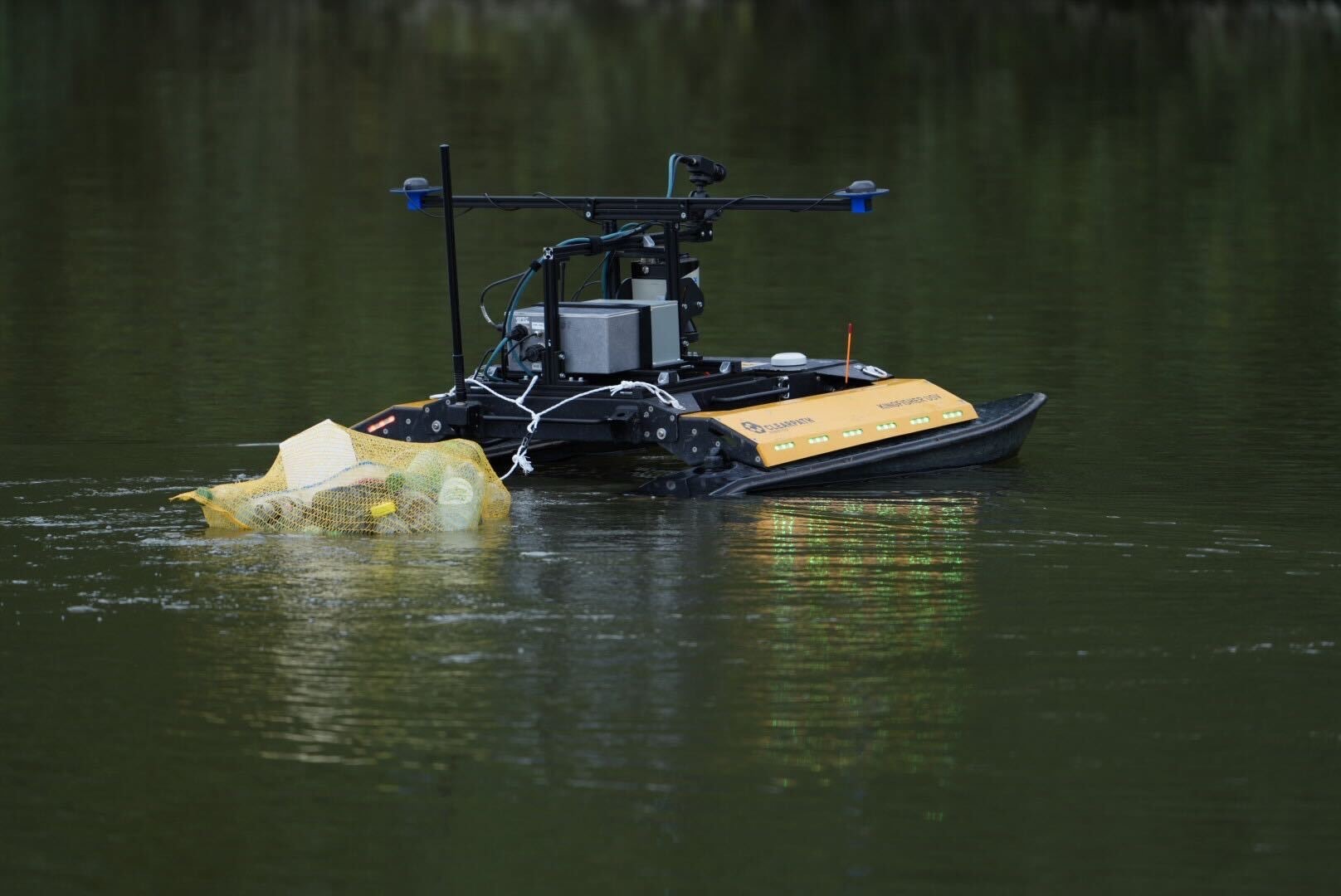}
        \caption{Research Platform}
        \label{fig:kingfisher}
    \end{subfigure}
    \caption{Examples of ASVs equipped with nets for capturing floating waste. The changing payload due to waste accumulation affects the ASV’s hydrodynamics and maneuverability.}
    \label{fig:capturing_waste}
        \vspace{-.5cm}
\end{figure}

Despite their potential, the effectiveness of ASVs for floating waste management can be limited by several complexities.
A particular challenge in capturing floating waste is the dynamic variation of the payload. As illustrated in \cref{fig:jellyfishbot}, one approach is to equip the ASV with a net to trap debris as it navigates the water. As waste accumulates in the net, the changing payload affects the ASV’s mass, moment of inertia, and drag characteristics, which in turn significantly alters its dynamics. Additionally, the ASV is exposed to external disturbances such as wind and water currents. These changes necessitate real-time adjustments to the control strategy to maintain accurate and stable operation during the waste collection process.

Traditional control techniques, such as Adaptive Control and Model Predictive Control (MPC), are widely used for ASV navigation. While effective in many applications, these methods face significant challenges, particularly in accurately estimating system dynamics, tuning parameters, and accounting for external disturbances. MPC, in particular, relies on precise models to optimize a cost function over a receding horizon, but precisely modeling the system dynamics for ASVs remains complex. 
This modeling is typically done through system identification (ARMA~\cite{ljung1998system}), or the model is learned with a Neural Network, as shown on an ASV by \cite{mahe_evaluation_2021}. The main drawback of such methods is that they rely on a model, and therefore are not designed to cope with a system whose dynamic model may change. 
Recent advances have introduced adaptive MPC techniques that account for payload variations, as demonstrated in the Roboat III project~\cite{wang_roboat_2023}. However, the described approach requires manual input for payload data, with automatic payload estimation planned for future work. 
To improve robustness, novel methods based on MPC estimate online either external perturbations, such as \cite{bahadorian2012robust}, or an unknown parameter of the model, such as \cite{zhang2021trajectory}, but they typically focus on a single variable. In contrast, our work addresses multiple dynamic changes simultaneously, including mass, inertia, drag, and environmental forces like wind and currents, to enhance the robustness of ASV control systems.

Simultaneously, Deep Reinforcement Learning (DRL) has recently emerged as a powerful tool in robotics, offering flexible and robust control strategies with growing applications in marine robotics and ASVs \cite{qiao2023survey}. For instance, \cite{zhao2020path} used a DRL-based controller for path following, demonstrating its effectiveness in managing ASVs within highly complex systems in simulation. Similarly, \cite{lin2023robust} introduced an IQN-based local path planner capable of handling unknown currents and avoiding obstacles. Various other studies have also explored RL techniques for ASV control problems \cite{zhang2021model, wang2020reinforcement, wang2022reinforcement}, though many of these focus predominantly on simulations, with limited real-world testing. 
In contrast, some research has emphasized evaluating DRL agents in real-world environments. For example, \cite{richard2022learning} trained a model-based RL agent for ASV trajectory following, demonstrating its robustness in simulation and its adaptability to real-world conditions in a zero-shot setup. Additionally, an RL-based approach combining guidance and heading control to improve tracking accuracy and controller robustness was proposed by \cite{wang2023path}. Wang et al. further showed that DRL outperformed Nonlinear-MPC in trajectory tracking, achieving lower tracking error and better disturbance rejection in river environments \cite{wang2023deep}.

Although these studies show promising results, they mainly focus on path-following tasks and are predominantly evaluated in simulation. While some agents are tested in real-world trials, their performance and robustness under varying conditions are not comprehensively analyzed.

In this work, we aim to explore and validate the robustness of an RL agent designed for capturing floating waste under different disturbance conditions. Our main contributions are:
\begin{itemize}
    \item Evaluation of the robustness of a DRL-based controller for ASVs through simulations and field tests, demonstrating its resilience against variations in center of mass, rotational drag, and external disturbances.
    \item Field experiments validating the DRL agent’s performance in realistic scenarios, showing minimal degradation compared to traditional controllers such as MPC. 
    \item Practical insights into the strengths and limitations of DRL-based control strategies alongside the code used to train the RL models, serving as a baseline for future ASV control research and development.
\end{itemize}

\section{Methodology}

\subsection{Problem Formulation}
\label{sec:methodo-problem}

The dynamic model of the ASV used in the simulation is based on Fossen’s six degrees of freedom model \cite{fossen2011handbook}, which governs the motion of the vehicle in response to internal and external forces:
\begin{equation}
    \mathbf{M}\dot\nu + \mathbf{D}(\nu)\nu + g(\eta) = \tau_{\text{thruster}} + \tau_{\text{disturbance}}
    \label{eq:fossen}
\end{equation}
Here, $\eta = [x, y, z, \phi, \theta, \psi]^T$ is the position vector, and $\nu = [u, v, w, p, q, r]^T$ is the velocity vector, composed of surge, sway, heave, roll, pitch, yaw. $\mathbf{M}$ represents the system inertia matrix and $\mathbf{D}(\nu)$ is the hydrodynamic damping matrix, given by:
\begin{equation}
    \mathbf{D}(\nu) = \mathbf{D}_l + \mathbf{D}_q(\nu)
    \label{eq:damping}
\end{equation}
It consists of linear ($\mathbf{D}_l$) and quadratic ($\mathbf{D}_q(\nu)$) terms, representing the linear and quadratic hydrodynamic damping coefficients in surge, sway, and rotation. Also in \cref{eq:fossen}, $g$ represents gravitational and buoyancy forces and moments. The terms $\tau_{\text{thruster}}$ and $\tau_{\text{disturbance}}$ refer to forces and moments generated by the thruster system and external disturbances \(e.g., wind, waves\), respectively. Given the ASV’s operating speed ($<1.5\ m/s$), the effects of Coriolis forces are considered negligible.

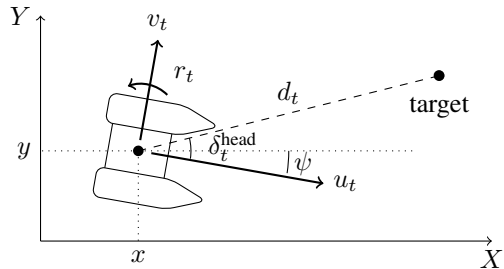
\begin{figure}[bth]
    \centering
    \usetikzlibrary{calc, angles, quotes}

\newcommand{\drawboat}[1][]{
    \begin{scope}[#1]
        \begin{scope}[shift={(-0.5,0)}] 
            \draw[rounded corners=1mm] 
            (-0.2,-0.5) -- (-0.2,0.5) -- (0,1) -- (0.2,0.5) -- (0.2,-0.5) -- cycle;
        \end{scope}
        \begin{scope}[shift={(0.5,0)}] 
            \draw[rounded corners=1mm] 
            (-0.2,-0.5) -- (-0.2,0.5) -- (0,1) -- (0.2,0.5) -- (0.2,-0.5) -- cycle;
        \end{scope}
        \begin{scope}[shift={(0,-0.4)}] 
            \draw[] (-0.3,0) -- (0.3,0);
        \end{scope}
        \begin{scope}[shift={(0,0.4)}] 
            \draw[] (-0.3,0) -- (0.3,0);
        \end{scope}
        \node at (0,0) (base_link) {$.$};
        \fill (0,0) circle (2pt); 

        \draw[->, thick] (base_link) -- ++(0,2.5) node[anchor=west] (u_arrow) {$u_{t}$};
        \draw[->, thick] (base_link) -- ++(-1.5,0) node[anchor=south] {$v_{t}$};
        
        \coordinate (v_tip) at ($(base_link)+(-0.8,0.25)$);
        \draw[->, thick] (v_tip) arc (145:210:0.5) node[pos=0.1, anchor=south west] {$r_{t}$};
    \end{scope}
}

\begin{tikzpicture}[scale=1.0] 
    \draw[->] (0,0) -- (6,0) node[anchor=north] {$X$};
    \draw[->] (0,0) -- (0,3) node[anchor=east] {$Y$};

    \drawboat[xshift=1.3cm, yshift=1.2cm, rotate=-100] 
    \coordinate (base_link) at (1.3,1.2); 
    \coordinate (T) at (5.3,2.2);
    \node at (T) {$\circ$};
    \fill (T) circle (2pt);
    \node[below of=T, node distance=12pt] {target}; 
    
    \draw[dotted] (base_link) -- (base_link |- 0,0) node [at end, below] {$x$};
    \draw[dotted] (base_link) -- (base_link -| 0,0) node [at end, left] {$y$};
    \draw[dotted] (base_link) -- (base_link -| 5,0) node [at end, right] {};
    
    \draw[dashed] (base_link) -- (T);
    \path (base_link) -- (T) node [midway, above] {$d_{t}$};

    \coordinate (u_tip) at ($(base_link)+(2.0,-0.3)$);
    \coordinate (y_ax) at ($(base_link)+(3.0,0,0)$);
    \pic [draw, -, "$\delta_{t}^\text{head}$", angle eccentricity=1.8, angle radius=0.7cm] {angle = u_tip--base_link--T};
    \pic [draw, -, "$\psi$", angle eccentricity=1.1, angle radius=2cm] {angle = u_tip--base_link--y_ax};

\end{tikzpicture}
    \caption{Definition of reference coordinate frame and target position observation in a 2D space.}
    \label{fig:dynamic_model_diagram}
        \vspace{-.5cm}
\end{figure}

This dynamic model is employed in simulation to train and evaluate an agent capable of autonomously controlling the ASV’s thrusters for the task of capturing floating waste. To perform this task, the ASV navigates over the waste, capturing it with a fixed net positioned between its hulls. The problem setup is illustrated in \cref{fig:dynamic_model_diagram}, where the target is specified in a two-dimensional space. The coordinates $x$ and $y$ represent the position of the ASV’s center, and $d_{t}$ indicates the distance to the goal. The task is considered successful when the ASV passes within $0.3\ \text{m}$ of the target’s position relative to its center. To focus solely on evaluating the capture task, the environment is kept free of obstacles. In anticipation of future integration with an onboard camera for perception, initial goal positions are constrained to a 90-degree field of view in front of the ASV, with an operational range of up to 10 meters.

\subsection{Platform}
\label{sec:platform}


To implement the RL-based control agent, we utilized a highly parallelized ASV simulation environment from  \cite{drl4asv}, which efficiently simulates buoyancy and hydrodynamics on a GPU. Similar to the setup in \cite{el2023rans}, this simulator is built on Nvidia's Isaac Sim and OmniIsaacGym \cite{makoviychuk2021isaac}, allowing for thousands of simulations to run in parallel. This massive parallelism accelerates training, enabling model-free methods like Proximal Policy Optimization (PPO)~\cite{schulman2017proximal} to quickly converge, making them competitive with traditional controllers like MPC or other optimal control methods.

For ROS integration and testing in a different simulation environment, we employed Gazebo with the UUV Simulation plugin \cite{bingham2019toward}. This setup provided an independent physics and hydrodynamics engine, which was valuable for an initial cross-validation of the performance of the RL agent and verifying for any overfitting to the specific physics implementation used in Isaac Sim. Additionally, Gazebo’s ROS integration facilitated seamless transitions to the real-world platform for testing.

In both simulation environments the hydrodynamic parameters used for the damping matrix (\cref{eq:damping}) were derived from system identification tests presented in \cite{drl4asv}. The values are:
$X_{u} = 0.00$,
$Y_{v} = 99.99$,
$N_{r} = 5.83$,
$X_{u|u|} = 17.26$,
$Y_{v|v|} = 99.99$,
$N_{r|r|} = 17.34$.

For the field tests, we used the Kingfisher ASV from Clearpath Robotics. This small catamaran-style vessel is $1.35\ \text{m}$ long, $0.98\ \text{m}$ wide, and weighs $35.82\ \text{kg}$. It is equipped with two independently controlled thrusters, one on each hull, with thrust commands ranging between $[-1.0, 1.0]$. A low-level proprietary microcontroller receives commands at a minimum frequency of $10\ \text{Hz}$ and forwards it to the motor controller. For onboard computation, an NVIDIA Jetson Xavier was used to run the RL agent. Power for the system was supplied by a pair of $22\ \text{Ah}$ 4-cell LiPo batteries. Although the ASV has cameras and a laser, they were not used; instead, goals were set programmatically. Localization was handled by a high-precision SBG Ellipse-D IMU and an RTK GPS with a dual-antenna GNSS receiver, providing position, velocity, and heading accuracy of $0.02\ \text{m}$, $0.03\ \text{m/s}$, and $0.5^{\circ}$, respectively.

\subsection{Control approach}

\subsubsection{RL}

Closely following the approach described in \cite{drl4asv}, the RL agent is based on the Proximal Policy Optimization (PPO) algorithm, using an actor-critic architecture with two hidden layers of 128 units each. This setup is designed for efficient control of the ASV’s thrusters in dynamic environments. The details of the observation space, action space, reward function, and training setup are presented in the following sections.

\textbf{Observation space: }
At each time step $t$, the observation space is defined as $\mathbf{o}_t \in \mathbb{R}^{6}$:

\begin{equation}
    \mathbf{o}_t = [u_{t} \quad v_{t} \quad r_{t} \quad \cos(\delta_t^{\text{head}}) \quad \sin(\delta_t^{\text{head}}) \quad d_{t}]^T
\end{equation}

The components consist of the surge speed $u_t$, sway speed $v_t$, and yaw rate $r_t$, which describe the linear and angular velocities of the ASV in the body frame. The orientation of the ASV relative to the goal is represented by $cos(\delta_t^{\text{head}})$ and $\sin(\delta_t^{\text{head}})$, providing a continuous representation of the bearing of the target. The final element $d_{t}$ denotes the distance to the goal at time step $t$ (\cref{fig:dynamic_model_diagram}).

\textbf{Action space: }
The action space, $\mathbf{u}_t = [\mathbf{u}_{t}^{\text{left}} \quad \mathbf{u}_{t}^{\text{right}}]^T$, consists of the command inputs for the left and right thrusters, respectively, at time step $t$. These inputs are transformed into thrust forces based on the characteristic force curves obtained from system identification experiments.

\textbf{Reward function: }
A compound reward function $r_t$ was designed to encourage the desired behaviors of the ASV agent during the capture task.
\begin{equation}
    r_t = r_t^\text{dist} + r_t^\text{head} + r_t^\text{energy} + r_t^\text{time} + r_t^\text{goal}
\end{equation}
Where each reward term is detailed as follows:
\begin{equation}
    r_t^\text{dist} = \lambda_{1} \cdot (d_{t-1} - d_{t})
\end{equation}
\begin{equation}
    r_t^\text{head} = \lambda_{2} \cdot ( e^{k_{1} (\delta_t^\text{head})})
\end{equation}
\begin{equation}
    r_t^\text{energy} =\lambda_{3} \cdot (e^{k_{2} E} - 1)
\end{equation}
\begin{equation}
    r_t^\text{time} = \lambda_{4}
\end{equation}
\begin{equation}
    r_t^\text{goal} =
    \begin{cases} 
        \lambda_{5}, & \text{if } d_{t} < d_{threshold} \\
        0, & \text{otherwise}
    \end{cases}
\end{equation}

$r_t^\text{head}$ encourages the ASV to align its heading with the goal direction.  
$r_t^\text{dist}$ encourages minimizing the distance to the goal, serving as an indicator of the ASV's advancement toward its target. 
$r_t^\text{energy}$ imposes a penalty on excessive actuation, promoting energy efficiency, where the energy consumption is defined as $E = (\mathbf{u}_{t}^{\text{left}})^2 + (\mathbf{u}_{t}^{\text{right}})^2$.  
$r_t^\text{time}$ discourages prolonged navigation by applying a step-wise penalty, motivating the ASV to reach the goal in minimal time.  
The $r_t^\text{goal}$ term provides a one-time reward upon successful goal attainment, with $d_{\text{threshold}}$ specifying the required proximity for completion.  
Each reward component is weighted by $\lambda_1, \lambda_2, \lambda_3, \lambda_4, \lambda_5$, while $k_1$ and $k_2$ are constants that fine-tune the sensitivity of the rewards to their respective errors.

\textbf{Training Details: }

\begin{table}[ht]
\centering
\caption{Parameter values for the reward function}
\renewcommand{\arraystretch}{1.3} 
\begin{tabular}{c c c c c c c}
\hline
Weights & $\lambda_1$ & $\lambda_2$ & $\lambda_3$ & $\lambda_4$ & $\lambda_5$ \\ 
Values & 1.0 & 1.0 & 0.001 & -1.0 & 10.0\\
\hline
Parameters & $k_1$ & $k_2$ & $d_{\text{threshold}}$ \\ 
Values & -4.0 & -0.1 & 0.1 m \\
\hline
\end{tabular}
\label{tab:reward_function_parameters}
\end{table}

The reward function parameters, detailed in Tab. \ref{tab:reward_function_parameters}, have been meticulously tuned to enhance the agent's performance. The cumulative reward per time step remains negative until the goal is reached, guiding the ASV's trajectory toward the target more effectively. A low value for $d_{\text{threshold}}$ was deliberately chosen to be lower than the actual distance required to capture the target. 

Tab. \ref{tab:rl_config} presents the key learning configurations used during the training, which was carried out on an Nvidia GeForce GTX 3090 GPU with 24GB RAM and was completed in approximately 10 minutes.

\begin{table}[!ht]\centering
\caption{Reinforcement learning configurations}\label{tab: }
\renewcommand{\arraystretch}{1.1} 
\scriptsize
\begin{tabular}{c c}\toprule
Parameters & Values \\\midrule
actor neural network & 128 $\times$ 128 \\
critic neural network & 128 $\times$ 128 \\
number of environments & 2048 \\
sample batch size & 16384 \\
max iterations & 1000 \\
max steps per episode & 3000 \\
time step size & 0.01 \\
\bottomrule
\end{tabular}
\label{tab:rl_config}
\end{table}

To train the DRL agent robustly, we applied domain randomization, introducing observation noise (position: $\pm3\ cm$, orientation: $\pm0.025\ rad$) and action perturbations to simulate sensor inaccuracies and control imprecision. A $\pm100\%$ variation was applied for $N_{r}$ due to its higher uncertainty, and $\pm10\%$ variation in the remaining damping parameters. A uniform randomization of the center of mass with up to $10\ cm$ around the center of mass was included to model payload changes. External disturbances were simulated using randomized forces ($\pm2.5\ N$) and torques ($\pm1.0\ N\cdot m$) applied to the center of mass of the ASV, challenging the agent without overwhelming it. We also applied up to 50\% randomization to thruster forces, accounting for battery fluctuations and eventual malfunctioning. Training followed a curriculum learning approach, gradually increasing difficulty over the first 300 epochs, then maintaining consistent conditions for the final 200 epochs to allow policy optimization. The agent’s final weights were used for the experiments.

\subsubsection{MPC}
To provide a baseline comparison, we implement an MPC controller for the USV.

The dynamic model of the ASV is a 2D simplification of Eq.~\ref{eq:fossen}, as proposed in \cite{wang_design_2018}, where only the linear term from the hydrodynamic damping matrix (Eq.~\ref{eq:damping}) is kept. This model is given by : 
\begin{equation}
    \mathbf{M}\dot\nu + \mathbf{D}_l\nu = \tau_{\text{thruster}}, \quad \text{where } \quad \nu = [u\quad v\quad r]^T 
\end{equation}
The kinematic equation relating velocities in the body and global frames is given by $\dot{\eta} = \mathbf{R}(\psi)\nu$, where $\mathbf{R}(\psi)$ is the rotation matrix from body frame to global frame and  $\eta = [x, y, \psi ]^T$ is the pose vector in the global frame (Fig.~\ref{fig:dynamic_model_diagram}).
 
Finally, the applied forces are described by
  \begin{equation}
      \tau_{\text{thruster}} = \mathbf{B}\mathbf{u} = 
      \begin{bmatrix}
      1 & 1 \\
      0 & 0 \\
      a/2 & a/2
    \end{bmatrix} 
    \begin{bmatrix}
      \mathbf{u}^{\text{left}} \\
      \mathbf{u}^{\text{right}} 
    \end{bmatrix} 
    \label{eq:forces}
     \end{equation}
All the other terms are defined in Section~\ref{sec:methodo-problem}.

From these equations, we can reformulate the dynamic model of the ASV:
\begin{equation}
    \dot{\mathbf{q}}(t) = f(\mathbf{q}(t),\mathbf{u}(t))
\end{equation}
Where $\mathbf{q} =[ x, y, \psi, u, v, r]^T$ is the state and $\mathbf{u}$ is the command (Eq.~\ref{eq:forces}).
The hydrodynamic parameters from $\mathbf{D}_l$ are identified with the system identification method described in \cref{sec:methodo-problem}. The mass $m$ is measured and the inertia $I_z$ is estimated from the weight and shape of the robot, allowing to estimate $\mathbf{M} = diag(m,m,I_z)$.

Finally, we formulate the optimal control problem (OCP) for the MPC as the least squares function that minimizes the deviations of state and control from the reference over the given time horizon. In this problem, the reference is simply the $(x,y)$ coordinates of the target.
The OCP is solved with the Acado~\cite{houska_acado_2011} framework. Following \cite{Falanga2018}, we execute one iteration of the optimization at each control loop, using as initial state the last available state, and the previous solution as initialization to the next optimization. That allows the optimization to run real-time, at the ASV control frequency (20Hz), with a 3 seconds time horizon.

\subsection{Evaluation Methodology}

The evaluation methodology focuses on comparing the performance of MPC and RL based controllers under varying disturbances.

The RL agent was trained in Isaac Sim, using domain randomization to simulate disturbances, with specific attention to rotational drag and the center of gravity (CoG) displacement with higher values of randomization. However, both the RL and MPC were evaluated under the same disturbance conditions in the Gazebo simulation environment, ensuring consistency in the physics simulation.

\begin{figure}
    \centering
    \includegraphics[width=0.9\linewidth]{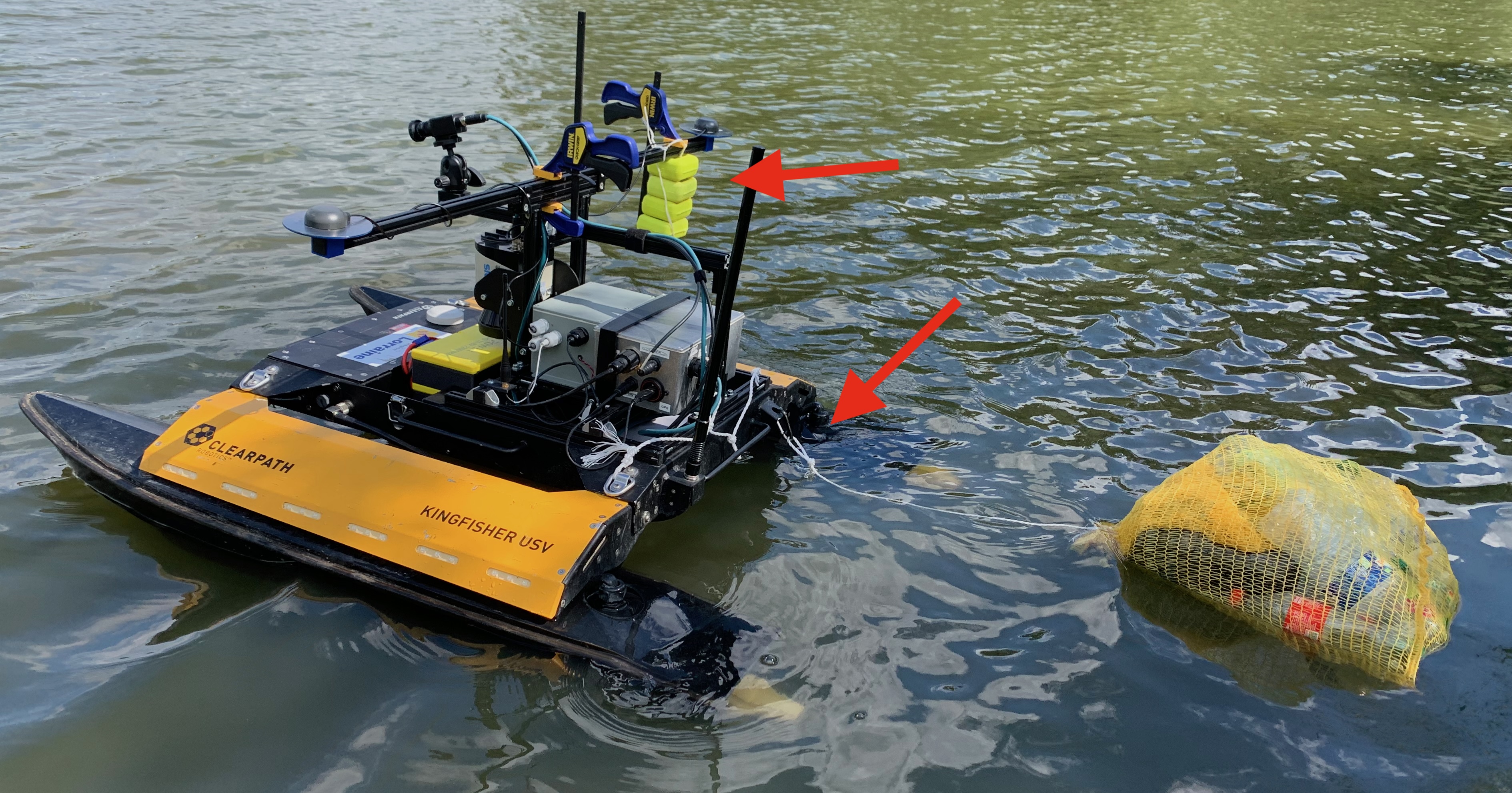}
    \caption{Disturbance applied during field tests: additional mass attached to the right of the center of mass and a floating net with water bottles creating asymmetric drag.}
    \label{fig:real-disturbances}
        \vspace{-.5cm}
\end{figure}

Field tests were conducted in real-world environments with natural disturbances, such as wind, as well as controlled disturbances. In the controlled tests, two additional disturbance scenarios were introduced: (1) a 4 kg weight was attached to the right of the center of mass, and (2) a floating net containing partially filled water bottles, creating a drag force with an offset $20\ cm$ from the center. These disturbances combined created asymmetric perturbations to the vessel and are shown in \cref{fig:real-disturbances}.

To assess performance, we measured three main metrics:  the duration to reach the goal normalized by the target distance, $T_{norm}$, the accumulated energy normalized by the target distance, $E_{\text{acc}}^{\text{norm}}$, and the increase in trajectory length compared to a straight line $\Delta_d$.
Given $d$, the distance to the goal when it is first received, $T$, the time to reach it, $d_{real}$ the length of the actual trajectory, $d_{threshold}$ the distance threshold below which the target is reached, and $N$ the total number of control actions until the goal is reached, the metrics are:
\begin{equation}
    T_{norm} = T/d
\end{equation}
\begin{equation}
    \Delta_d = d_{real} - (d - d_{threshold})
\end{equation}
\begin{equation}\label{eq:accenergy}
    E_{\text{acc}}^{\text{norm}} = {(\sum_{i=1}^{N}(|\mathbf{u}_{t}^{\text{left}}|+|\mathbf{u}_{t}^{\text{right}}|))/d} 
    \end{equation}

Robustness was evaluated by comparing each controller's performance in undisturbed and disturbed conditions, with MPC being compared against disturbed MPC, and RL against disturbed RL. This approach ensures a fair comparison between each controller's robustness to disturbances as their objective functions and optimization processes differ.

\section{Experiments}

\subsection{Simulation}

The simulation in Gazebo was used to assess the agent’s robustness to variations in key parameters, including surge drag ($X_{u}$), rotational drag ($N_{r}$), mass ($M$), center of mass ($CoM$), and moment of inertia about the Z-axis ($I_{z}$). Among these, we selected for detailed analysis the two with the most significant impact on the agent’s performance during preliminary investigation $CoM$ and $N_{r}$. The $CoM$ shifts from 0.0 to 12.5 cm along the Y-axis, and $N_{r}$ from 0 to 20.

No additional disturbances (e.g., wind or currents) were applied in the simulation to isolate and evaluate the effects of the parameter variations. 
To conduct our evaluation, we compare the performances of the agents with a set of goal and initial speed conditions, when we vary either $CoM$ or $N_{r}$.
For a given value of $CoM$ and $N_{r}$, the  goal distances were set at 1-meter intervals between 3 and 9 meters, with bearing angles ranging from -45 to 45 degrees in 5-degree increments. Also, we tested three different initial speeds (0, 0.5, and 1 m/s), resulting in a total of 399 goals per environment.

\begin{figure}
    \centering
    \includegraphics[width=0.99\linewidth]{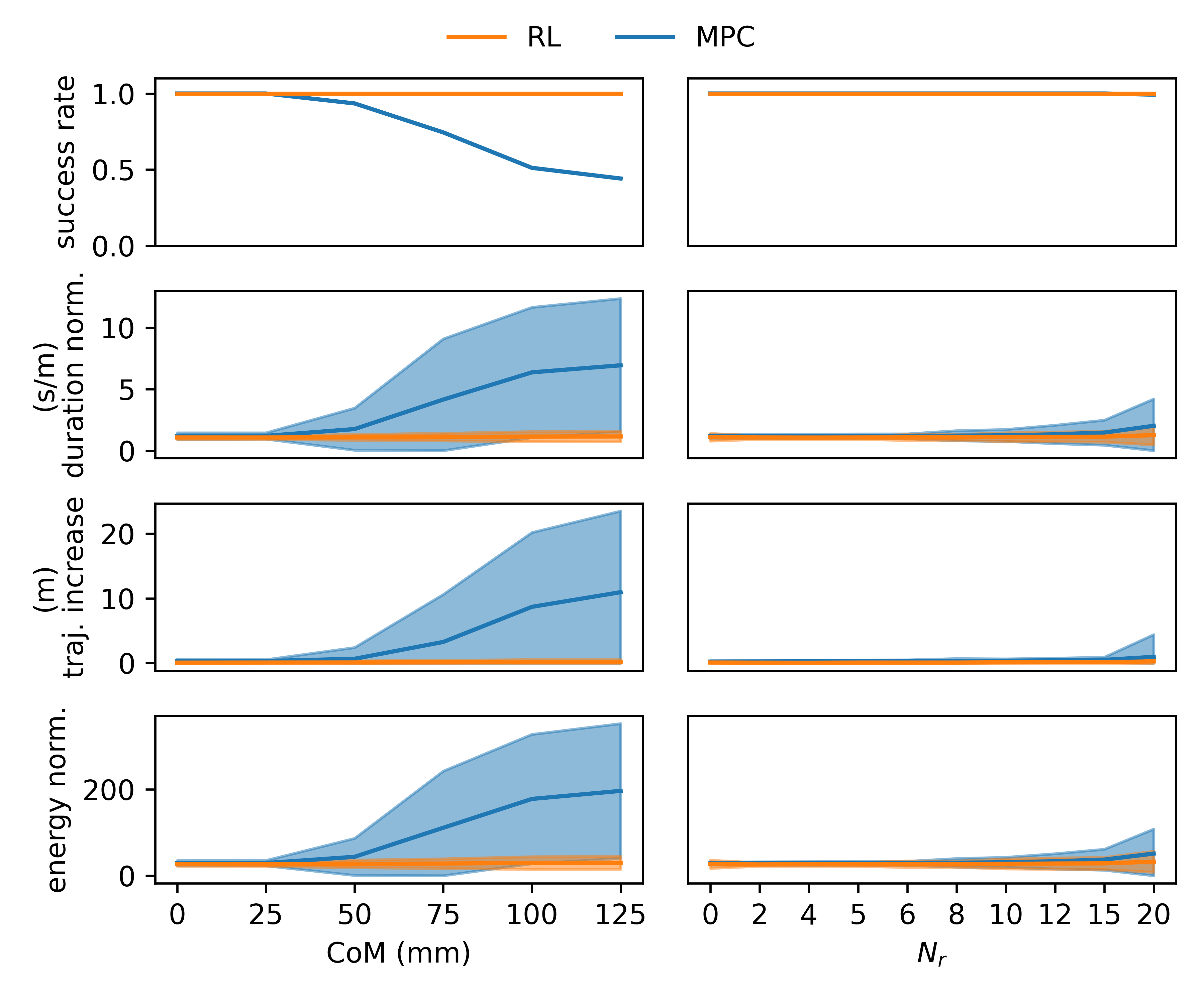}
    \caption{Impact of changes in center of mass and rotational drag on agent performance in simulation. Mean and standard deviation of success rate, normalized duration, $T_{norm}$, and trajectory increase, $\Delta_d$, and normalized accumulated energy, $E_{\text{acc}}^\text{norm}$, are shown.}
    \label{fig:sim-overview}
        \vspace{-.5cm}
\end{figure}

The mean and standard deviation for each experiment are shown in \cref{fig:sim-overview} for the three metrics, normalized duration $T_{norm}$, trajectory increase $\Delta_d$ and normalized accumulated energy, $E_{\text{acc}}^\text{norm}$. The figure also shows the success rate. It is possible to observe that the RL agent maintained a consistent success rate near 100\% across all variations in the center of mass ($CoM$) and rotational drag ($N_r$), whereas the MPC controller experienced a significant drop in performance as $CoM$ shifted.

As the $CoM$ shifted, the MPC’s success rate declined sharply, and all three metrics grew substantially, indicating its sensitivity to these changes. In contrast, the RL agent exhibited robustness across all conditions showing minimal changes in normalized duration and accumulated energy, and experienced negligible trajectory increase, even under large $CoM$ shifts and variations in $N_r$.

\begin{figure}
    \centering
    \includegraphics[width=0.99\linewidth]{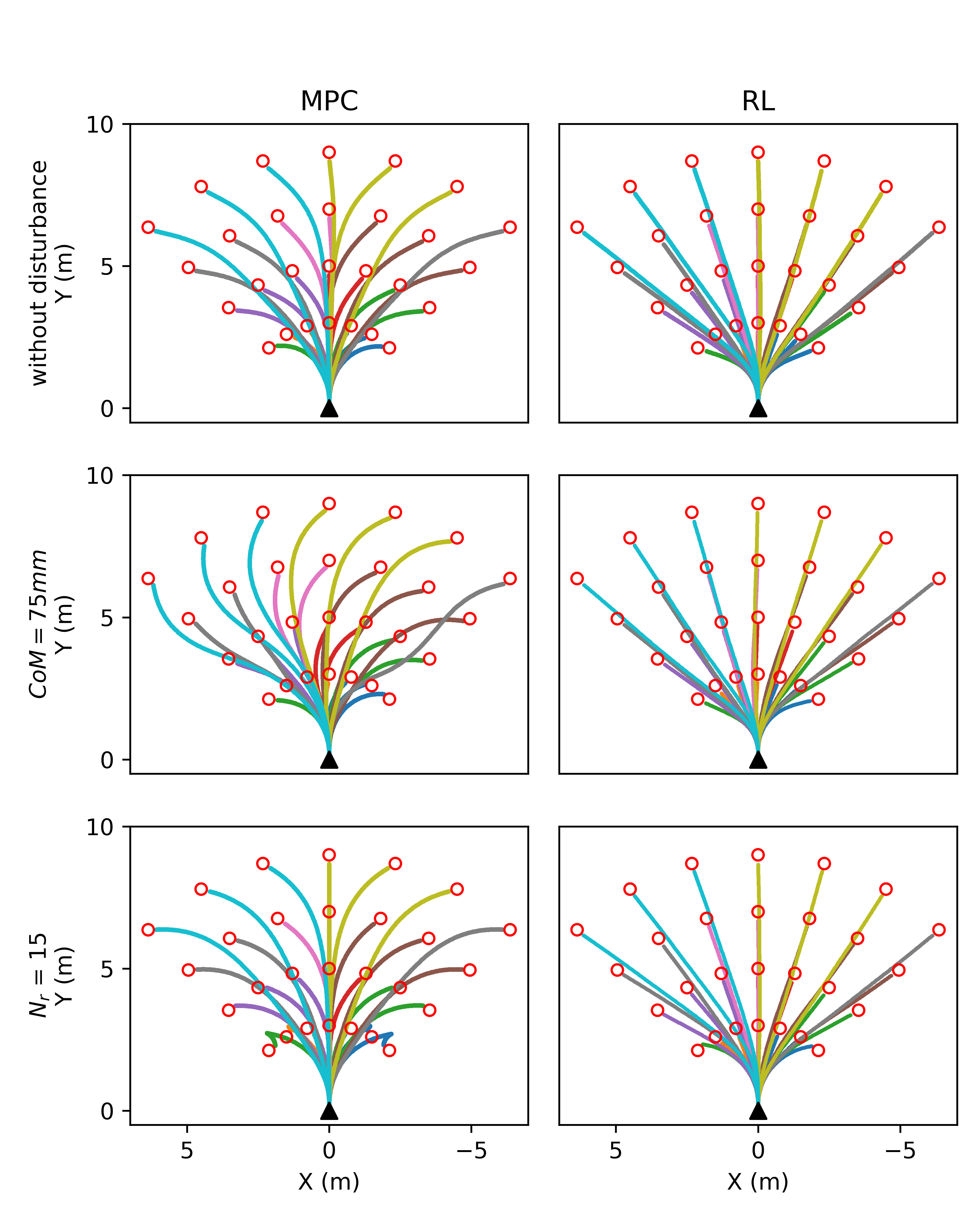}
    \caption{Trajectories for a subset goals (red circles) in simulation under varying disturbances. The RL agent maintains consistent trajectories, while the MPC shows varied deviation patterns as disturbances grow.}
    \label{fig:sim-trajectories}
        \vspace{-.5cm}
\end{figure}

The impact of the disturbances on the trajectories is shown in \cref{fig:sim-trajectories}, which displays three scenarios: normal conditions, $CoM = 75\ \text{mm}$, and $N_r = 20$. The RL agent consistently maintains a straight trajectory despite varying disturbance conditions. Notably, the $CoM$ disturbance causes a distorted trajectory, while the high $N_r$ disturbance forces the MPC agent to overshoot and backtrack, particularly at the 3 m and 45-degree goals.

\subsection{Field tests}

We present the results of field tests conducted under two conditions: (i) a baseline scenario, referred to as \textit{normal}, where the boat operated solely under natural weather disturbances, and (ii) a perturbed scenario, referred to as \textit{disturbed}, in which additional mass and drag were applied, as depicted in \cref{fig:real-disturbances}. In the disturbed test, the $4\ kg$ mass was attached with an offset of approximately $38\ cm$ to the right-side, changing the ASV's center of mass, and a floating net with water bottles partially filled with water (weighing $2.8\ kg$ outside of water) was attached behind the ASV to simulate additional drag. The attachment point, as indicated in \cref{fig:real-disturbances}, was offset by approximately $21\ cm$ to the right. These offsets combined, produced an asymmetric disturbance, causing a counterclockwise torque to the Z axis of the vessel as it accelerates forward. 

We quantified the weather disturbances, mostly wind, by recording RTK IMU measurements over 8 minutes without input commands. The ASV reached angular speeds of $-0.14$ to $0.07$ rad/s, surge speeds of $-0.15$ to $0.34$ m/s, and sway speeds of $-0.13$ to $0.05$ m/s. These disturbances are significant compared to the ASV’s maximum velocity of $1.6$ m/s, highlighting the impact of environmental forces on performance.

\begin{figure}
    \centering
    \includegraphics[width=0.99\linewidth]{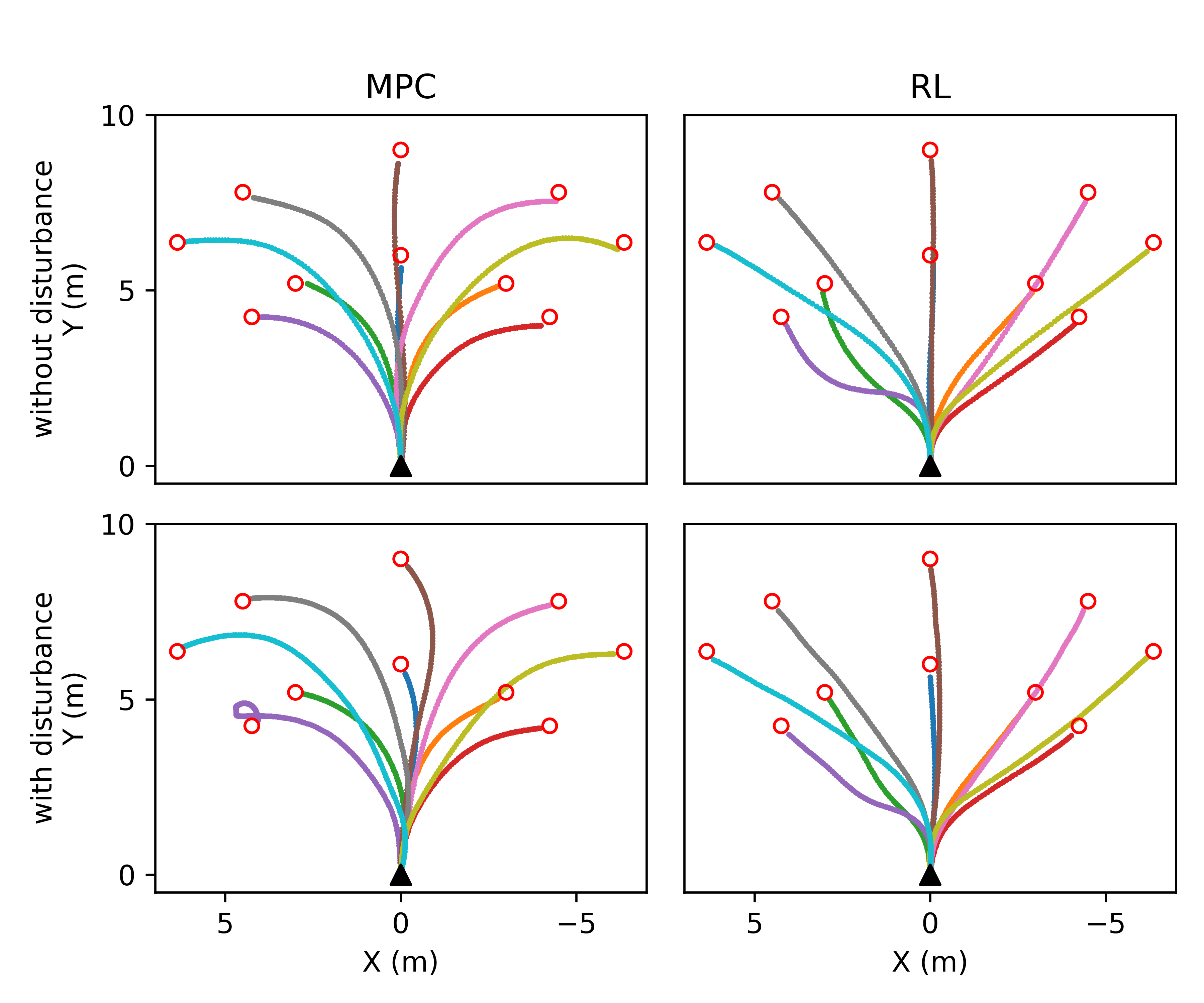}
    \caption{Field test trajectories for the ASV under \textit{normal} and \textit{disturbed} conditions. While RL agent mostly retains trajectory stability, the MPC shows tendency for deviations aligned to the direction of the external disturbances.}
    \label{fig:field-trajectories}
    \vspace{-.5cm}
\end{figure}

The field test trajectories in  \cref{fig:field-trajectories}, show the RL agent maintaining a straight path, while MPC reacts slowly to disturbances, similar to simulation.

Finally, \cref{tab:field-tests-metrics} highlights the robustness of the agents to the disturbances. The table displays the mean value of each metric, across all goals.
It shows that the degradation in normalized time, $T_{norm}$, is similar for both agents, although slightly better for the RL. Nonetheless, the mean value of $T_{norm}$ remains better for the RL agent when compared to the MPC, both in \textit{normal} and \textit{disturbed}.
It is also worth noting that the RL agent shows an improvement, i.e negative degradation, in the trajectory deviation metric ($\Delta_d$). The reason is that the drag, added in \textit{disturbed}, slows the vessel, allowing it to initiate earlier rotations toward the target, thereby reducing overall trajectory deviation. Nonetheless, the drawback is the increase in energy consumption ($E_{\text{acc}}$) which is likely due to its active counteraction of torque and drag forces to maintain straight trajectories.

\begin{table}[!ht]
    \centering
    \scriptsize
    \begin{tabular}{lr|ccc}\toprule
    Agent & Metric & Normal & Disturbed & Degradation \% \\\midrule
    \multirow{3}{*}{RL} 
    & $T_{norm}$ & 1.094 & 1.247 & 13.96 \\
    & $E_{\text{acc}}$ & 28.619 & 36.248 & 26.66 \\
    & $\Delta_d$ & 0.246 & 0.160 & -34.88 \\\midrule
    \multirow{3}{*}{MPC} 
    & $T_{norm}$ & 1.175 & 1.350 & 14.89 \\
    & $E_{\text{acc}}$ & 33.457 & 36.457 & 8.96 \\
    & $\Delta_d$ & 0.697 & 0.775 & 11.23 \\
    \bottomrule
    \end{tabular}
    \caption{Robustness evaluation: Field Tests Metrics for RL and MPC under \textit{normal} and \textit{disturbed} Conditions}
    \label{tab:field-tests-metrics}
        \vspace{-.5cm}
\end{table}

\section{Discussion}

Our results demonstrate that the RL agent effectively reacted to various sources of disturbances, maintaining straight trajectories toward the goal without explicitly modeling or directly observing these disturbances. This outcome suggests that domain randomization, combined with a well-designed simulation environment, can be efficient for real-world ASV applications, particularly in scenarios where disturbances are complex or difficult to quantify.

Nonetheless, achieving robust performance required iterative refinement of the simulation environment and field validation. The primary challenges centered on reducing the simulation-to-reality gap and carefully shaping the reward function to guide the agent toward desired behaviors. Inaccuracies in the simulation model or improper reward settings often led the RL agent to develop unexpected behavior. 

On the contrary, the primary advantage for the MPC is its structured approach and explainability. For example, by tuning some coefficients from the model (in particular $N_r$), we were able to achieve the best results for the \textit{normal} condition experiments presented in this paper. 

\begin{figure}
    \centering
    \includegraphics[width=0.9\linewidth]{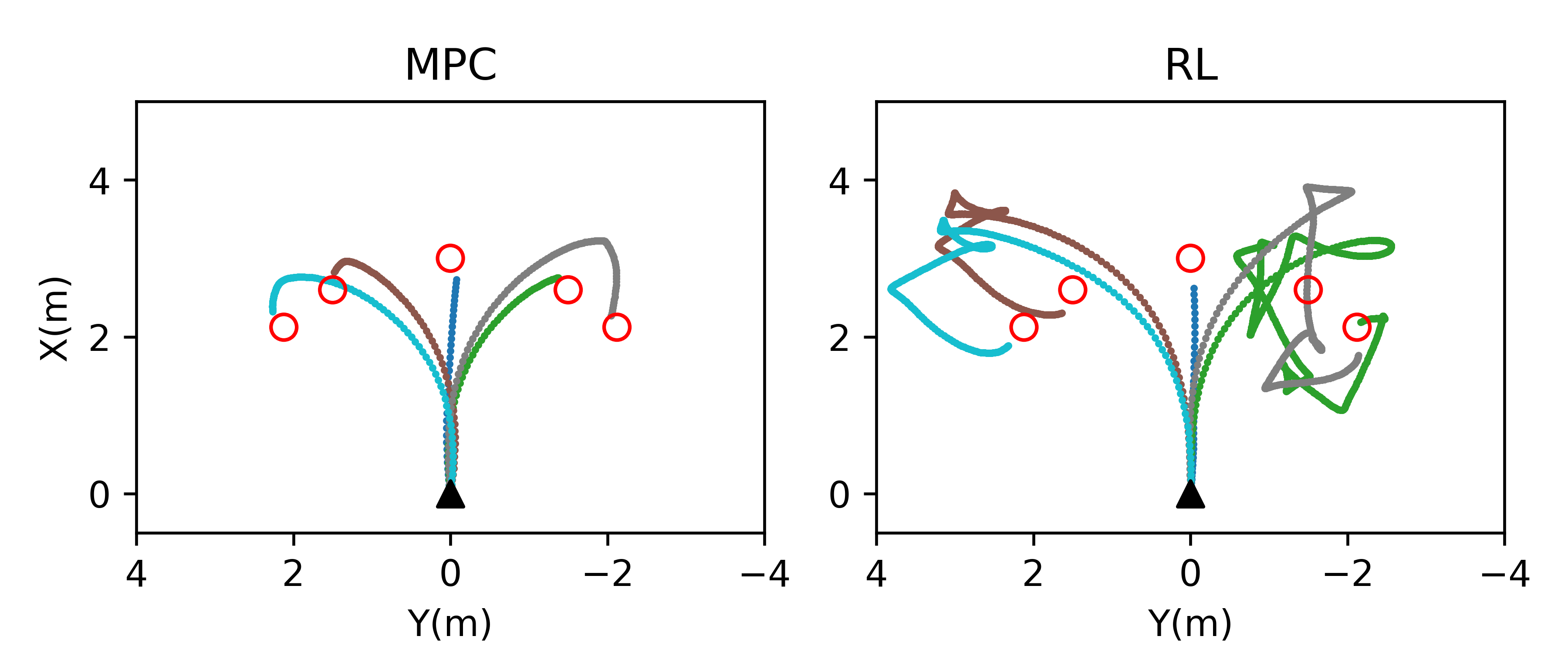}
    \caption{Illustration of limitations encountered during field tests. The RL agent struggles with in-place rotations when approaching nearby goals with high bearing angles.}
    \label{fig:limitations}
        \vspace{-.5cm}
\end{figure}

Finally, the primary limitation observed for the RL agent in field tests was its difficulty to perform in-place rotations, which are necessary for reaching nearby goals with high bearing angles (\cref{fig:limitations}). These maneuvers pose significant challenges even for human operators due to the ASV's underactuated and non-holonomic dynamics. However, we observed that the RL policy struggled with in-place rotations significantly more in the field than in simulation. Upon careful analysis, we attribute this discrepancy to differences between the real and simulated actuators, which are further discussed in \cref{sec:challenges}. This mismatch caused the RL agent to overshoot in-place rotations, ultimately preventing the agent from efficiently reaching the target. In contrast, the MPC adopted a more conservative strategy, either executing slower rotations to align with the target or reversing to reach the desired position.

\section{Challenges and Lessons Learned}
\label{sec:challenges}

Conducting field experiments with the Kingfisher ASV presented several challenges, primarily related to hardware limitations, control delays, and environmental disturbances. Addressing these issues was essential to ensure reliable data collection and validate our experimental results.

\begin{figure}[bt]
    \centering
    \begin{subfigure}{0.49\linewidth}
        \includegraphics[width=\linewidth]{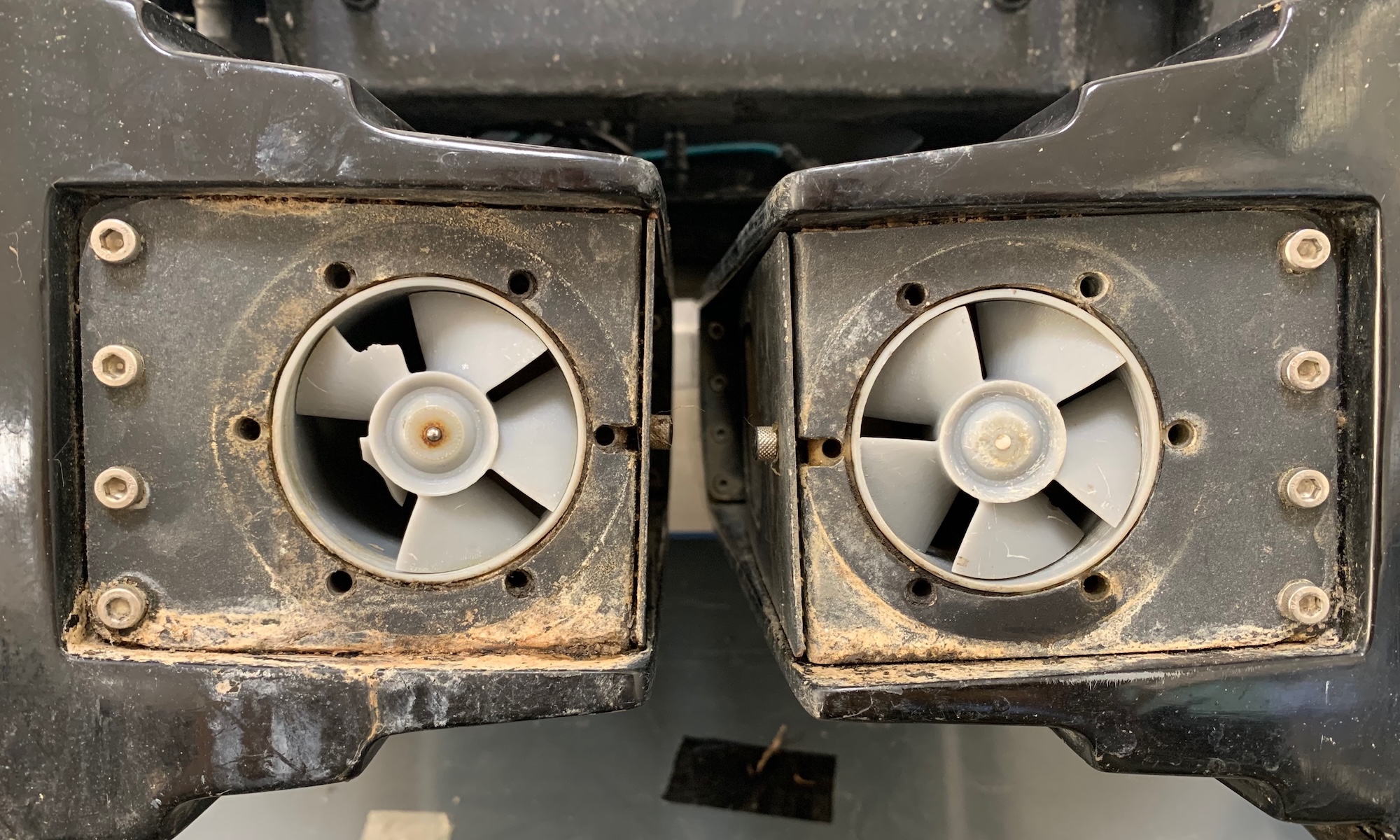}
        \caption{Broken original propeller}
        \label{fig:broken_propeller}
    \end{subfigure}
    \begin{subfigure}{0.49\linewidth}
        \includegraphics[width=\linewidth]{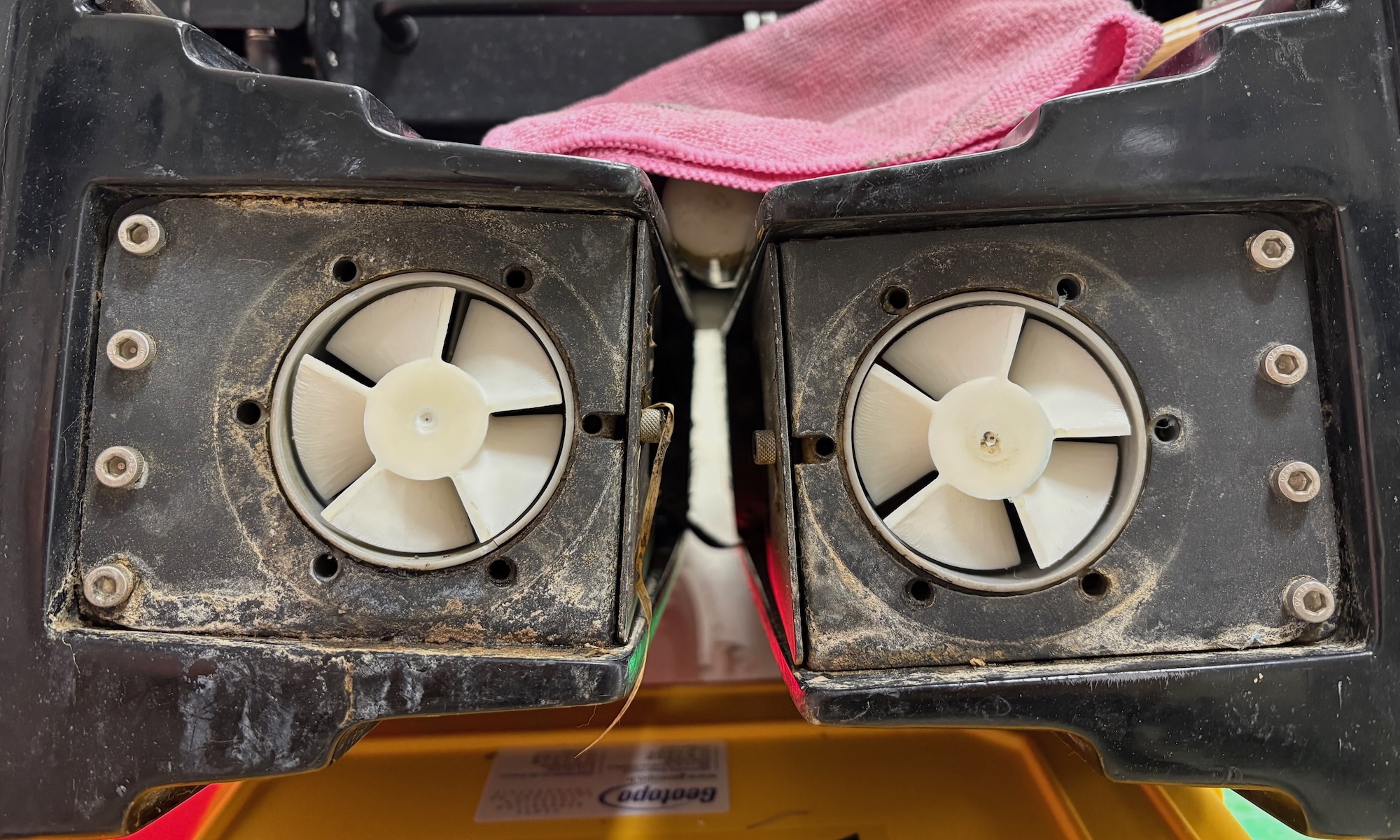}
        \caption{3D-printed replacement}
        \label{fig:fixed_propeller}
    \end{subfigure}
    \caption{The replacement propeller required redesign for feasibility}
    \label{fig:propeller}
    \vspace{-.5cm}
\end{figure}

One illustrative issue we encountered was a set of broken propeller blades, as shown in \cref{fig:broken_propeller}. The Kingfisher ASV we used is an aging platform that Clearpath discontinued over a decade ago. Although it remains operational and robust, regular maintenance is essential to ensure reliable performance. Having the necessary tools and technical expertise for field repairs is indispensable when working with such platforms. In this particular case, replacing the damaged propellers required repeating system identification experiments to update the propeller characteristic force curve, ensuring consistency with our simulation environment. When planning ASV experiments, it is important to allocate repair and maintenance time into the schedule and consider the robot's lifespan when referencing benchmark studies, such as \cite{sears2024otterrospickingprogramminguncrewed}.

One of the most significant challenges we encountered was diagnosing the limitation shown in \cref{fig:limitations}. Specifically, discrepancies between simulation and field tests raised questions about whether in-place rotation issues stemmed from inaccuracies in hydrodynamic modeling, ASV inertia, or thruster response characteristics. Since these factors are interrelated, isolating their individual contributions based solely on field-collected data was challenging.

\begin{figure}[tb]
    \centering
    \includegraphics[width=0.99\linewidth]{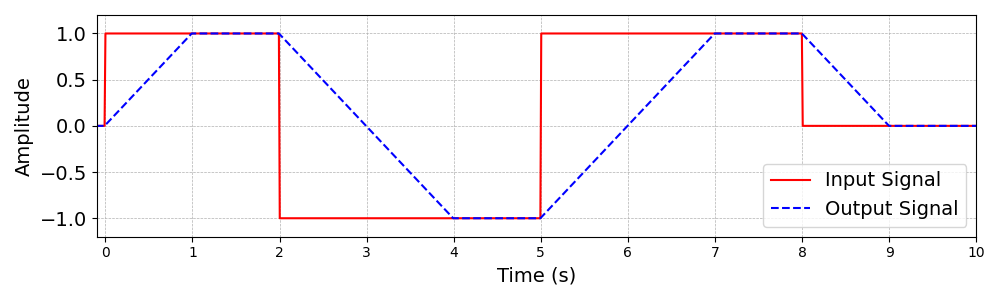}
    \caption{Motor command input response observed on the microcontroller. The output sent to the motor controller presents a significant delay.}
    \label{fig:cmd-drive-response}
    \vspace{-.5cm}
\end{figure}

The audible response of the propellers to manual commands indicated a potential actuation delay. This prompted further investigation into the proprietary microcontroller responsible for converting input commands into Pulse Position Modulation (PPM) signals for the motor controllers. Using an oscilloscope, we measured the signal propagation and precisely analyzed the commands received by the motor controllers. Our findings revealed that the applied command is rate-limited, meaning that each update is constrained by a maximum allowable change from the previous command. This rate limitation introduced a cumulative delay, resulting in a total response time of approximately 2 seconds when changing the input from -1.0 to 1.0, as shown in \cref{fig:cmd-drive-response}. Understanding this delay is crucial for better modeling, particularly when implementing controllers that require precise actuation timing.

Environmental factors also played a significant role in system performance. One challenge was quantifying the weather disturbances affecting the ASV without specialized sensors such as an anemometer. Initially, we relied on local weather forecasts for wind conditions. However, this approach proved impractical, as localized wind effects in the test area often deviated from forecasted conditions, and the unknown aerodynamics of the ASV made it difficult to directly relate wind velocity to the forces and torques acting on the vessel. To address this, we adopted a simple yet effective alternative: before initiating new experiments, we allowed the ASV to drift freely while recording its position. By analyzing data from the onboard RTK-GPS and IMU, we estimated the accelerations and peak velocities induced by external disturbances. This method provided a practical way to quantify environmental forces acting on the vessel and assess the robustness of the RL policy in real-world conditions.

\section{Future Prospects}


The first necessary modification is updating the thruster model in simulation to accurately replicate the delay described in \cref{sec:challenges}. Incorporating this rate-limited response into the simulator will improve the fidelity of our model, allowing for better alignment between simulation and real-world performance. We expect this adjustment to significantly mitigate the overshooting issues observed in \cref{fig:limitations}, leading to improved control stability and accuracy.


Beyond refining the simulation model, we aim to enhance the ASV’s autonomous capabilities by incorporating additional control tasks. Velocity and trajectory tracking, as well as station-keeping, are fundamental for a wide range of ASV applications and will contribute to more reliable operation in dynamic environments. Furthermore, integrating obstacle avoidance will reduce the need for constant human supervision, making deployments more efficient and scalable.


To achieve fully autonomous floating waste collection, we plan to integrate a camera-based perception system capable of detecting and classifying objects such as plastic bottles. This perception layer will identify relevant targets and generate goal positions for the reinforcement learning-based agent controlling the capture task. By enabling real-time perception and decision-making, this enhancement will allow the ASV to autonomously navigate and collect waste, improving its effectiveness in environmental cleanup missions.

\section{Conclusion}

This work evaluated the robustness of a DRL-based agent for controlling ASVs in complex and disturbed environments. 
We incorporated domain randomization during training and conducted extensive evaluation in simulation, where the DRL agent demonstrated consistent performance and resilience against variations in mass, rotational drag, and environmental disturbances, maintaining trajectory stability. We further confirm the results with real-world field tests.
Our results underscore the potential of DRL for ASV control, offering a flexible and robust alternative to traditional methods. Based on the experience gained during our field tests, future work will focus on improving simulation environments to better capture real-world dynamics and expanding the agent’s capabilities to handle broader operational scenarios. Our open source implementation aims to facilitate further research and development in this field, contributing to more efficient and adaptive ASV navigation solutions.


\section*{ACKNOWLEDGMENT}
This research was supported by the French Agence Nationale de la Recherche, under grant ANR-23-CE23-0030 (project R3AMA).

\bibliographystyle{IEEEtran}
\bibliography{IEEEabrv,references}

\end{document}